\documentclass[letterpaper, 10pt, conference]{ieeeconf}
\IEEEoverridecommandlockouts
\usepackage{amsmath, amssymb, amsfonts}
\usepackage{mathtools}
\usepackage{cite}
\usepackage[font=small, labelfont=bf]{caption}
\usepackage[ruled, lined, linesnumbered, commentsnumbered, longend, noend]{algorithm2e}
\usepackage[table]{xcolor}
\usepackage{algorithmic}
\usepackage{graphicx}
\usepackage{textcomp}
\usepackage{multirow}
\usepackage{makecell}
\usepackage{diagbox}
\usepackage{nicefrac}
\usepackage{booktabs}
\usepackage{breqn}
\usepackage{gensymb}
\usepackage{cuted}
\usepackage{capt-of}
\usepackage{siunitx}
\usepackage{pifont}

\makeatletter
\let\NAT@parse\undefined
\makeatother
\usepackage[bookmarks=false,linkcolor=blue, urlcolor=blue, citecolor=blue]{hyperref}
\hypersetup{
    colorlinks=true,
    linkcolor=red,
    filecolor=magenta,      
    urlcolor=black,
    pdfstartview={FitH},
    citecolor=blue
}

\newcommand{\systemName}{HumiTransfer}
\newcommand{\datasetName}{\texttt{HumiMimic}}

\title{\LARGE \bf \datasetName: Mimetic Embodiment Transfer of Human–Humanoid Interaction from In-the-Wild Videos} 

\author{
Nghia Le$^{1}$, 
Khang Nguyen$^{2}$,
Ba Long Tran$^{3}$,
Thi Quynh Nhu Pham$^{1}$,
Binh X. Nguyen$^{1}$,
Tuong Do$^{1,4,5}$, \\
Erman Tjiputra$^{1}$,
Hoan Nguyen$^{6}$,
Tung D. Ta$^{7}$,
Baoru Huang$^{4}$,
Hai-Nguyen Nguyen$^{3}$, 
Anh Nguyen$^{4}$
}

\begin{document}

\maketitle 

\begin{strip}
    \vspace{-33pt}
    \centering
    \includegraphics[width=1.00\linewidth]{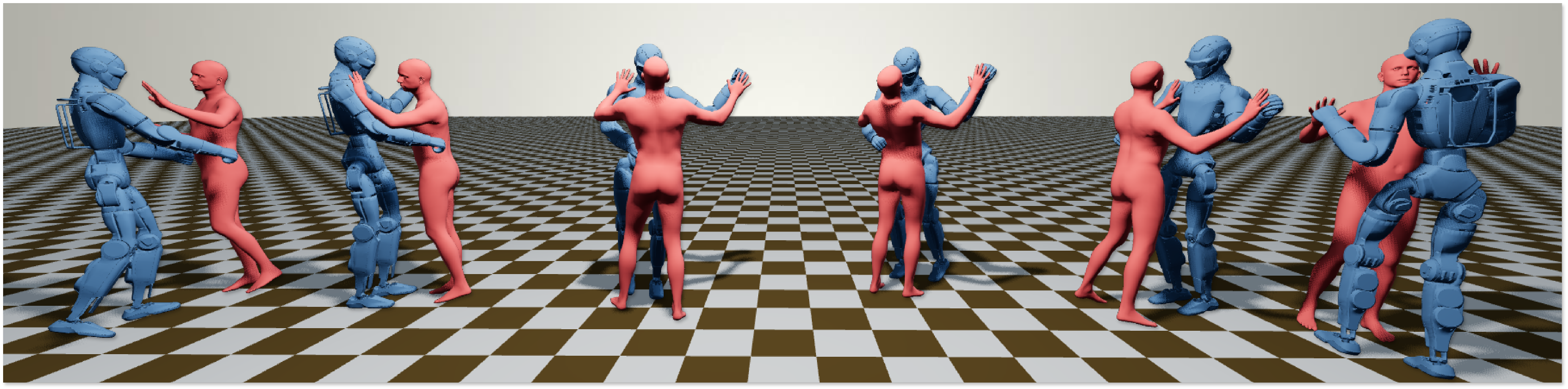}
    \vspace{-16pt}
    \captionof{figure}{\textbf{Overview of~\systemName:} Given an in-the-wild human–human interaction video, our mimetic embodiment transfer replaces one participant with a humanoid while preserving the interaction semantics and dynamics that characterize the original human behavior. Through this,~\systemName~ensures that the resulting behavior is physically plausible and compatible with different humanoid embodiments.}
    \vspace{-15pt}
    \label{fig:teaser}
\end{strip}

\begin{abstract}
    Enabling a humanoid to interact naturally with human partners requires preserving not only individual motions but also the spatial, semantic, and contact relationships underlying the interaction (\textit{i.e.}, hugging, dancing, hand-shaking, etc.). Furthermore, a practical solution should generalize across various humanoid embodiments with distinct morphologies and kinematic constraints. To tackle this challenge, we propose~\systemName, a mimetic embodiment transfer framework that converts in-the-wild human-human interaction videos into physically plausible human-humanoid interactions.~\systemName~first reconstructs 3D human motion with consistent foot-ground contact, then transfers one participant to a target humanoid embodiment via interaction modeling, using weighted joint-relative directions and anatomical offsets to prioritize interaction-critical inter-body relationships, while post-retargeting optimization enforces balance, contact, and collision avoidance. Subsequently, we apply~\systemName~to $1,200$ in-the-wild videos, producing approximately $500,000$ frames of diverse interactions for 
    two humanoid robots, resulting in our~\datasetName~dataset. Across both robots,~\systemName~achieves superior motion fidelity, preserves inter-body relationships, and maintains contact accuracy compared with existing baselines.  Moreover, we demonstrate that~\datasetName~supports motion generation methods for synthesizing interactive humanoid motions following a human partner. Finally, our real robot deployments confirm the physical executability of the transferred motions in real-world settings. Our project is available at \href{https://aioznetwork.github.io/HumiMimic/}{https://aioznetwork.github.io/HumiMimic/}.
\end{abstract}
\renewcommand{\thefootnote}{}
\setcounter{footnote}{0}
\footnotetext{

$^{1}$ AIOZ Ltd., Singapore

$^{2}$ MBZUAI, Abu Dhabi, United Arab Emirates

$^{3}$ RMIT University, Vietnam Campus

$^{4}$ University of Liverpool, UK

$^{5}$ National Tsing Hua University, Taiwan

$^{6}$ University of Information Technology, VNUHCM, Vietnam.

$^{7}$ Keio University, Mita, Minato-ku, Tokyo, Japan
}

\vspace{-2pt}
\section{Introduction}
\label{sec:introduction}
Humans are adept at participating in the social world not only through explicit instruction but also by observing and responding to others' behaviors~\cite{bandura1961transmission}. Such a capacity is commonly referred to as \textit{mimesis}: the representation and communication of experience through interaction in both semantic and dynamic manners~\cite{inamura2004embodied, nakamura2007mimetic}. Indeed, interaction cannot be acquired by imitating isolated motions alone. Meaningful human-to-human interactions fundamentally arise from the evolving relationship between participants. For example, making a humanoid dance with another human requires not merely reproducing the learned whole-body motions, but also being aware of interpersonal spacing and orientation, synchronizing their rhythm and movement phase, coordinating contact, and continually adapting to the partner's motion despite variations in embodiment. Herein, mimesis offers a natural account of how observed human behavior can translate into actions for humanoid robots, connecting motion recognition, motion generation, and embodied communication, thus opening up a vast range of applications in human-humanoid interactions.

\begin{table*}[t]
    \centering
    \vspace{4pt}
    \caption{\textbf{Key Attributes of Interaction Datasets:} Comparisons of key attributes between~\datasetName~dataset and existing human-motion and humanoid-motion datasets in terms of collection setting, scale, action diversity, and supported humanoid embodiments.}
    \vspace{-4pt}
    \resizebox{1.00\linewidth}{!}{
        \begin{tabular}{r|c|c|c|c|c}
            \toprule
            \diagbox{Dataset}{Attribute} & \makecell{No. Subjects and\\Means of Collection} & \makecell{No. Sequences} & \makecell{Type of Actions} & \makecell{Multi-Person\\Interaction} & \makecell{Robot/Humanoid\\Embodiment} \\
            \midrule \midrule
            AMASS~\cite{mahmood2019amass}
            & 344, MoCap & 11,265 & Daily/locomotion & No & No \\
            Human3.6M~\cite{ionescu2013human3}
            & 11, MoCap & $\sim$3.6M frames & 15 daily & No & No \\
            KIT Whole-Body~\cite{mandery2015kit}
            & 238, MoCap & 2,972 & Whole-body everyday & Some & No \\
            3DPW~\cite{von20183dpw}
            & 7, IMU + single-view RGB & 60 clips & Walking, interacting & Some & No \\
            Harmony4D~\cite{khirodkar2024harmony4d}
            & 24, multi-view RGB & 208 & Close interactions & Yes & No \\
            CHICO~\cite{sampieri2022poseCHICO}
            & 20, multi-view RGB & 240 clips & 7 industrial interactions & No & KUKA \\
            \midrule
            H2O~\cite{he2024learning}
            & AMASS & $\sim$10k & Whole-body everyday & No & Unitree H1 \\
            AgiBot World~\cite{bu2025agibot}
            & Teleop & $\sim$3000 hours & Dexterous/loco-manip & No & AgiBot G1 \\
            CLONED~\cite{li2025clone}
            & AMASS + MoCap & 345 & Whole-body loco-manip & Some & Unitree H1 \\
            OmniH2O-6~\cite{he2024omnih2o}
            & Teleop & $\sim$72k frames & 6 everyday tasks & Some & Unitree H1 \\
            \midrule
            \rowcolor[HTML]{EFEFEF}
            \textbf{\datasetName~(Ours)} & Multiple (in-the-wild) & 500K frames & Diverse (sports, daily, social) & Yes & Unitree R1 Edu \& HighTorque Mini-hi \\
            \bottomrule
        \end{tabular}
    }
    \vspace{-10pt}
    \label{tab:dataset_comparison}
\end{table*}

In practice, existing humanoid motion datasets~\cite{he2024learning, he2024omnih2o, jiang2025kaiwu} are predominantly collected through high-fidelity teleoperation or motion-capture systems in controlled environments. Although these pipelines provide accurate, robot-compatible motions, they are often tied to specific embodiments and capture only a subset of interaction behaviors within limited environments. Advances in scene-aware and occlusion-robust human pose estimation~\cite{luo2022embodied, cheng2019occlusion}, mesh-based body fitting~\cite{bogo2016keep}, and learning-based monocular reconstruction~\cite{shin2024wham} have made it possible to recover whole-body human motion from videos. Despite promising results, they often overlook human-human interactions due to the scarcity of suitable datasets. Publicly available videos, by contrast, contain abundant and diverse human-human interactions captured in real-world settings, offering a scalable source for deriving human-humanoid interaction motions.

Recent advances in whole-body motion learning~\cite{jiang2024harmon, nguyen2025td, nguyen2025doublyaware} have expanded the range of human-like behaviors that humanoids can synthesize. Still, learning behaviors such as human-to-human interactions remain fundamentally constrained by the availability and diversity of training data: a humanoid must not only reproduce an individual motion but also coordinate with its human partner~\cite{wang2026magic}. Meanwhile, retargeting methods offer a means of converting observed human behavior into embodiment-specific compatibility, yet existing approaches~\cite{joao2025gmr, huang2026human2humanoid, ProtoMotions} primarily optimize single-agent motion, neither preserving the inter-body relationships underlying the interaction nor identifying which semantic cues to optimize across embodiments. Therefore, the central challenge of mimetic embodiment transfer is to translate one participant's motion into a humanoid embodiment while preserving the interaction semantics with its human partner.

To tackle this challenge, we propose~\systemName, a new mimetic embodiment transfer framework that translates in-the-wild human-human interactions into human-humanoid interactions, as shown in Fig.~\ref{fig:teaser}.~\systemName~takes various interaction sequences as video frames to extract human motion, then transfers one human participant's poses to two distinct humanoid embodiments: the $23$-DoF HighTorque Mini-hi and the $26$-DoF Unitree R1 Edu, resulting in our~\datasetName~dataset. We also benchmark existing retargeting methods to showcase the diversity and complexity of interactions in~\datasetName. In summary, our main contributions are twofold as follows:
\begin{itemize}
    \item We propose~\systemName, a mimetic embodiment transfer framework that translates one human into a humanoid while preserving the semantics and distances underlying the interaction and satisfying embodiment-specific kinematic and physical constraints.
    \item We introduce~\datasetName, a scalable embodied dataset that contains $1,200$ human-humanoid interaction sequences (${\sim} 500,000$ frames) by extracting body frames from in-the-wild human-human interaction videos. Our code and dataset will be made publicly available.
\end{itemize}

\vspace{-4pt}
\section{Related Work}
\label{sec:related_work}

\noindent \textbf{Human Motion Datasets.} 
Human motion datasets have traditionally relied on motion capture systems in controlled environments~\cite{joo2015panoptic, trumble2017total}. Human3.6M~\cite{ionescu2013human3} and CMU MoCap~\cite{cmu_mocap} provide accurate 3D poses, while AMASS~\cite{mahmood2019amass} unifies diverse MoCap sources in the SMPL representation~\cite{loper2015smpl}. Multi-person interaction datasets, such as CHI3D~\cite{fieraru2021three} and Harmony4D~\cite{khirodkar2024harmony4d}, introduce physical interactions, while 3DPW~\cite{von20183dpw} and KIT Whole-Body~\cite{mandery2015kit} incorporate more unconstrained settings. Recent humanoid datasets, H2O~\cite{he2024learning}, AgiBot World~\cite{bu2025agibot}, CLONED~\cite{li2025clone}, and OmniH2O-6~\cite{he2024omnih2o}, provide humanoid-compatible motion but are primarily collected through teleoperation or controlled data acquisition. In general, existing datasets lack the joint provision of large-scale, in-the-wild multi-person interactions and humanoid-compatible motion. To fill this gap, this work introduces~\datasetName, a dataset of human-humanoid interactions featuring high-fidelity, physically plausible poses/motions.

\noindent \textbf{Human-to-Humanoid Retargeting.} 
Human-to-humanoid retargeting ranges from inverse kinematics and optimization-based methods~\cite{aristidou2011fabrik, joao2025gmr} to learning-based approaches~\cite{choi2021self, peng2018deepmimic}. Optimization-based approaches primarily preserve the motion of the transferred agent through joint or end-effector alignment~\cite{tang2024humanmimic, he2024learning, luo2023perpetual}. Uniform transfer~\cite{ProtoMotions} preserves relative geometry through scaling, yet it is sensitive to embodiment differences. Proportional transfer~\cite{joao2025gmr, he2024learning} adapts motion to the target embodiment but introduces distortions due to morphological dissimilarities. Under the umbrella of interactive robotics, these methods fall short of determining the relational cues critical to interaction, making it difficult to maintain high-fidelity, embodiment-specific interactions. Meanwhile, physics-based approaches improve the geometric quality of interactions~\cite{pan2025spider}, where recent interaction-aware approaches~\cite{huang2026human2humanoid, huang2026learning} explicitly consider relationships between interacting bodies, while graphics-oriented methods~\cite{cheynel2025reconform, jang2024geometry} preserve interaction semantics. Within this research scope,~\systemName~extends optimization-based human-to-humanoid retargeting, enabling embodiment-specific interaction with another human participant in in-the-wild settings.

To recap, Table~\ref{tab:dataset_comparison} summarizes our method with existing human-motion and humanoid datasets, highlighting its unique combination of scale, in-the-wild multi-person interactions, and support for diverse humanoid embodiments.


\begin{figure*}[t]
    \centering
    \vspace{4pt}
    \includegraphics[width=1.00\linewidth]{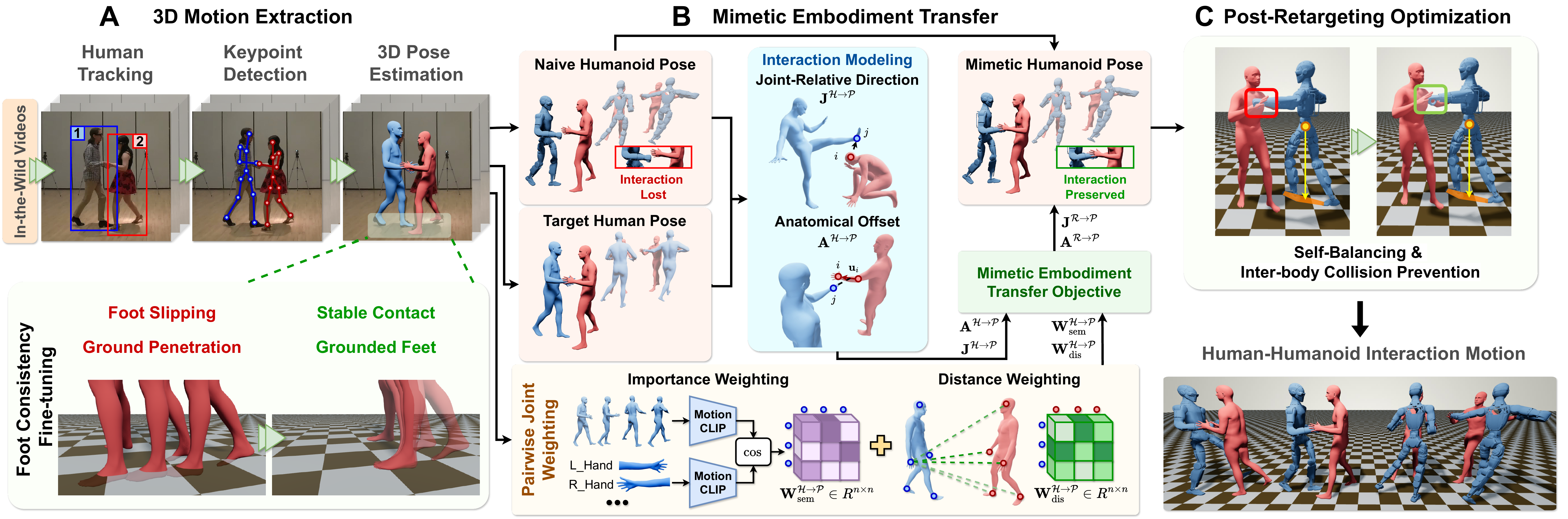}
    \vspace{-16pt}
    \caption{\textbf{Overview of~\systemName:} \textbf{(A)} Given an in-the-wild video of a human-human interaction,~\systemName~tracks human participants, detects body keypoints, reconstructs mesh-based full-body frames, and refines the resulting frames to ensure consistent ground-foot contact (Sec.~\ref{sec:h2i_extraction}). \textbf{(B)} Next, the poses of one participant are initialized on the target humanoid through constrained inverse kinematics.~\systemName~then represents the interaction semantics using joint-relative directions $\mathbf{J}^{\mathcal{R} \rightarrow \mathcal{P}}_t$ and anatomical offsets $\mathbf{A}^{\mathcal{R} \rightarrow \mathcal{P}}_t$. Within this process, semantic and distance-based weights, $\mathbf{W}^{\mathcal{H} \rightarrow \mathcal{P}}_{\mathrm{sem}}$ and $\mathbf{W}^{\mathcal{H} \rightarrow \mathcal{P}}_{\mathrm{dis}, t}$, are used to prioritize the inter-body relationships that are important to the observed interaction, allowing their alignment across human and humanoid embodiments (Sec.~\ref{sec:met_transfer}). \textbf{(C)} The retargeted pose is further refined to maintain humanoid balance and suppress unintended inter-body collisions while preserving intended contacts and interaction dynamics, yielding the final human-humanoid interaction (Sec.~\ref {sec:post_retargeting}). By applying~\systemName~onto the Unitree R1 Edu and the HighTorque Mini-hi, we obtain~\datasetName, our human-humanoid interaction dataset.}
    \vspace{-18pt}
    \label{fig:dataflow}
\end{figure*}

\section{\systemName: Mimetic Embodiment Transfer for Human-Humanoid Interaction}
\label{sec:met_h2i}
As shown in Fig.~\ref{fig:dataflow}, we introduce a three-stage framework: (\textit{i}) a framework for human–humanoid interaction with reconstructed 3D interaction motions; (\textit{ii})~\systemName, a mimetic embodiment transfer for interaction-preserving human-to-humanoid motion retargeting; and (\textit{iii}) a post-retargeting optimization stage. After which, we obtain~\datasetName, our human-humanoid interaction dataset.

\subsection{Human-Humanoid Interaction Data}
\label{sec:h2i_extraction}

\subsubsection{Data Collection \& Statistics}
\begin{figure}[h]
    \centering
    \vspace{-5pt}
    \includegraphics[width=1.00\linewidth]{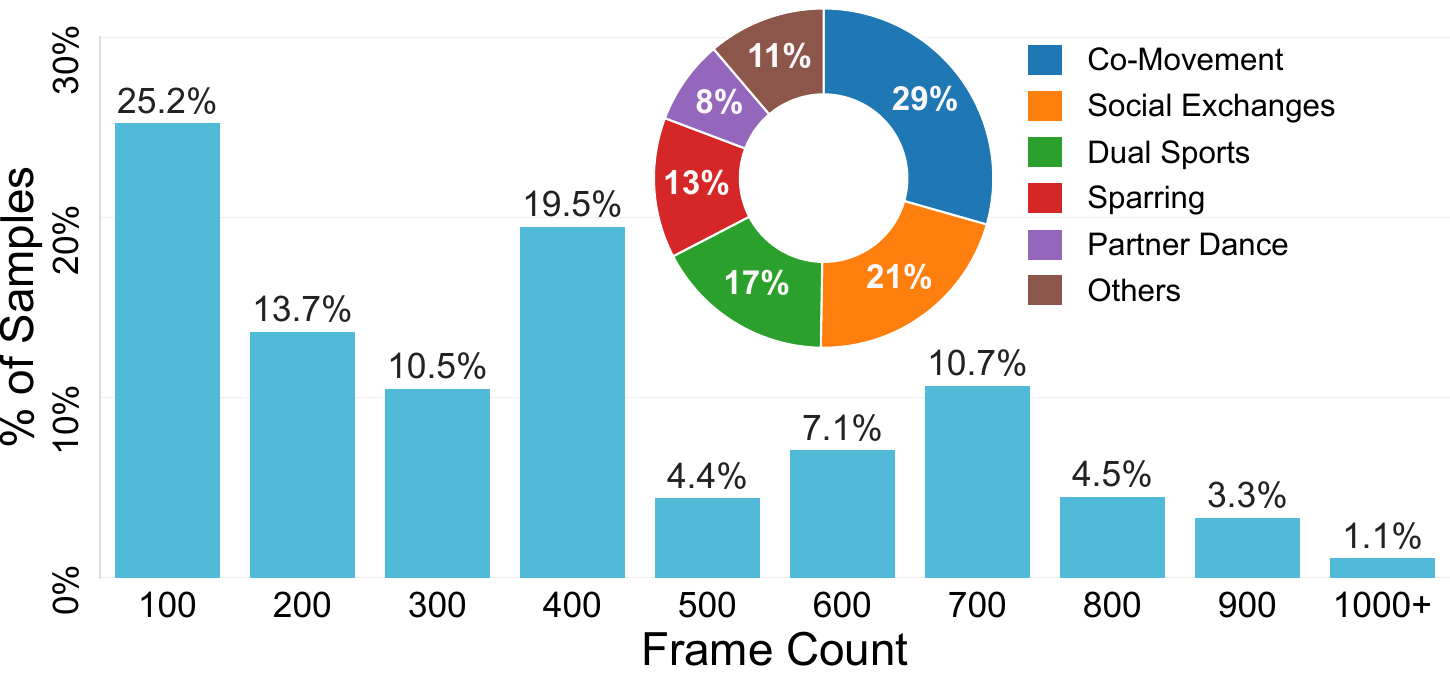}
    \vspace{-14pt}
    \caption{\textbf{Dataset Statistics:} Distribution of interaction lengths (in frames) and interaction diversity (in percentages) from the collected in-the-wild human-human interaction videos.}
    \vspace{-6pt}
    \label{fig:vidnoid_stats}
\end{figure}

We gather high-quality videos of in-the-wild human-human interactions from YouTube, TikTok, Facebook, and other Internet sources, involving only two participant subjects. As depicted in Fig.~\ref{fig:vidnoid_stats}, our resulting dataset contains $1,200$ two-person interaction videos, yielding about $500,000$ processed frames (approximately $4.6$ video hours). Each sequence ranges from $100$ to $1,000$ frames, with more than half exceeding $300$ frames. Regarding diversity, our dataset covers a broad spectrum of interactions, such as co-movement ($29$\%), social exchanges ($21$\%), dual sports ($17$\%), and sparring ($13$\%). 

\subsubsection{Extraction of Human-Human Interaction Motion}
As we noticed image-quality distortions, camera instability, and scene complexity in the collected in-the-wild videos, we perform an automatic filtering step~\cite{salazar2026xmopipe} to remove low-fidelity frames and re-encapsulate high-fidelity ones, retaining approximately $73\%$ of all frames. Next, in each re-encapsulated interaction sequence, we employ a multi-person tracker~\cite{sun2022_dancetrack} to detect humans in each frame using bounding boxes. For each bounding box, a pre-trained keypoint detector~\cite{yin2024whac} is used to detect the human's key positions within the bounding box, which are subsequently used to perform SMPL reconstruction~\cite{pavlakos2019expressive}. Thus, to ensure the physical plausibility of extracted poses, we optimize two complementary objectives in each video: (a) whole-body shape-fitting optimization and (b) foot-ground contact consistency fine-tuning.

\paragraph{Whole-Body Shape Fitting}
The pose parameter ($\alpha$) and the body shape parameter ($\beta$) of the SMPL model~\cite{pavlakos2019expressive} $\mathcal{H}_{\mathrm{SMPL}}$ are obtained by minimizing:
\begin{equation}
    E_{\mathrm{shape}} = E_{j} + \lambda_{\alpha} E_{\alpha} + \lambda_{\beta} E_{\beta}
    \label{eq:shape_fitting}
\end{equation}
where $E_j$ represents the re-projection error between the detected and projected body joints, $E_{\alpha}$ and $E_{\beta}$ are pose and shape prior errors~\cite{pavlakos2019expressive}; each is weighted with $\lambda_{\alpha}$ and $\lambda_{\beta}$.

\paragraph{Foot-Ground Contact Consistency Fine-tuning}
To further enhance the quality of the reconstructed human model (Eq.~\ref{eq:shape_fitting}), we minimize the contact error between the fitted SMPL and the ground, computed as the foot consistency error and a Zero Moment Point (ZMP) regularizer~\cite{kajita2003biped}:
\begin{equation}
    E_{\mathrm{contact}} = E_{\mathrm{consist}} + \lambda_{\mathrm{zmp}} \, E_{\mathrm{zmp}},
    \label{eq:foot_ground_contact}
\end{equation}
with the foot consistency error $E_{\mathrm{consist}}$ formulated as:
\begin{equation}
    E_{\mathrm{consist}} = \sum_{i \in \mathcal{F}} \mathbb{I}_{i,t} \left[ \left \| J_{i, t} - J_{i, t-1} \right\| + \left\|(J_{i, t} - \mathbf f)^\top\mathbf n^\ast \right\|^2 \right],
    \label{eq:foot_consistent_error}
\end{equation}
where $\mathcal{F} \subset \mathcal{H}_{\mathrm{SMPL}}$ denotes the set of foot joints of the SMPL model, $\mathbb{I}_{i,t}$ denotes the ground-contact indicator function for foot joint $i$, $J_{i ,t}$ denotes the position of joint $i$ derived from its forward kinematics at time $t$, $\mathbf{f}$ is the reference foot position, and $\mathbf{n}^{*}$ is the estimated normal of the planar ground. Meanwhile, the second term in Eq.~\ref{eq:foot_ground_contact} enforces the mesh-based human body's balance while maintaining consistent motion with the ground. 

By applying Eq.~\ref{eq:shape_fitting} and Eq.~\ref{eq:foot_ground_contact} to all detected humans, we obtain sets of SMPL models that serve as the frames $\{ \mathcal{H}_{\mathrm{SMPL}} \}_{t=0}^T$, forming our dataset for retargeting a humanoid's motions interacting with a human partner.

\vspace{-4pt}
\subsection{Interaction Modeling \& Mimetic Embodiment Transfer}
\label{sec:met_transfer}
Given the extracted sequence of human body frames $\{ \mathcal{H}_{\mathrm{SMPL}, t} \}_{t=0}^T$ and a humanoid embodiment $\mathcal{R}$, our goal is to transfer this sequence into a humanoid's motion while preserving interaction semantics and geometric relationships. 

\subsubsection{Humanoid Poses Warm-Starting}
As a warm start, we initialize humanoid poses $\{ \mathcal{R}_t \}_{t=0}^{T}$ by retargeting the human body frames $\{ \mathcal{H}_{\mathrm{SMPL}, t} \}_{t=0}^{T}$ through inverse kinematics, denoted as $E_{\mathrm{IK}}$, similar to~\cite{joao2025gmr, huang2026learning}. Specifically, at each discrete timestep $t$, the human pose $\mathcal{H}_{\mathrm{SMPL},t}$ is scaled according to the humanoid's proportion. The humanoid embodiment $\mathcal{R}$ is then initialized in its default state; thus, its joint poses are optimized to match the scaled target frames while respecting the kinematic constraints and joint limits.

\subsubsection{Modeling of Human-Humanoid Interaction}
To enable mimetic embodiment transfer for human-humanoid interaction, we model these interactions as two matrices: the joint-relative direction matrix and the anatomical offset matrix between the interacting subjects' joints.

\paragraph{Joint-Relative Direction Matrix} 
Let $J_{i, t}^{\mathcal{H}}$, $J_{i, t}^{\mathcal{R}}$, $J_{i, t}^{\mathcal{P}}$ be the $i$-th joint in the human body frame $\mathcal{H}_{\mathrm{SMPL}}$, the humanoid embodiment $\mathcal{R}$, and the interacting human partner $\mathcal{P}$ at the discrete timestep $t$. We, therefore, introduce the joint-relative directions of the humanoid $\mathcal{R}$ and its human partner $\mathcal{P}$, written as an $n \times n$ square matrix:
\begin{equation}
    \mathbf{J}^{\mathcal{R} \rightarrow \mathcal{P}}_t = 
    \begin{bmatrix}
        r^{\mathcal{R} \rightarrow \mathcal{P}}_{11, t} & r^{\mathcal{R} \rightarrow \mathcal{P}}_{12, t} & \dots & r^{\mathcal{R} \rightarrow \mathcal{P}}_{1n, t} \\
        \vdots & \vdots & \ddots & \vdots \\
        r^{\mathcal{R} \rightarrow \mathcal{P}}_{n1, t} & r^{\mathcal{R} \rightarrow \mathcal{P}}_{n2, t} & \dots & r^{\mathcal{R} \rightarrow \mathcal{P}}_{nn, t} \\
    \end{bmatrix},
    \label{eq:joint_relative_direction_matrix}
\end{equation}
where $\mathcal{R} \rightarrow \mathcal{P}$ reads as the relation of $\mathcal{R}$ to $\mathcal{P}$, and each entry in $\mathbf{J}^{\mathcal{R} \rightarrow \mathcal{P}}_t$ is computed as $r^{\mathcal{R} \rightarrow \mathcal{P}}_{ij, t} =  J_{i, t}^{\mathcal{R}} - J_{j, t}^{\mathcal{P}}$. Eq.~\ref{eq:joint_relative_direction_matrix} encodes the joint-based directional offset vectors between two interacting subjects. Similarly, $\mathbf{J}^{\mathcal{H} \rightarrow \mathcal{P}}_t$ can be formulated as the joint-relative direction matrix between the human body frame $\mathcal{H}_{\mathrm{SMPL}}$ and the human partner $\mathcal{P}$.

\paragraph{Anatomical Offset Matrix} 
To reinforce the naturalness of the interaction, we also incorporate the matrix encoding anatomical offsets among the subjects $\mathcal{R}$ and $\mathcal{P}$ at each discrete timestep $t$. Its $n \times n$ square matrix is:
\begin{equation}
    \mathbf{A}^{\mathcal{R} \rightarrow \mathcal{P}}_t = 
    \begin{bmatrix}
        a^{\mathcal{R} \rightarrow \mathcal{P}}_{11, t} & a^{\mathcal{R} \rightarrow \mathcal{P}}_{12, t} & \dots & a^{\mathcal{R} \rightarrow \mathcal{P}}_{1n, t} \\
        \vdots & \vdots & \ddots & \vdots \\
        a^{\mathcal{R} \rightarrow \mathcal{P}}_{n1, t} & a^{\mathcal{R} \rightarrow \mathcal{P}}_{n2, t} & \dots & a^{\mathcal{R} \rightarrow \mathcal{P}}_{nn, t} \\
    \end{bmatrix},
    \label{eq:anatomical_offset_matrix}
\end{equation}
where $a^{\mathcal{R} \rightarrow \mathcal{P}}_{ij, t} = \left( \mathbf{u}_{i,t}^{\mathcal{R}} \right)^\top \left( J_{i, t}^{\mathcal{R}} - J_{j, t}^{\mathcal{P}} \right)$ with $\mathbf{u}_{i,t}^{\mathcal{R}}$ is the unit vector from the $i$-th joint to its immediate parent joint. Eq.~\ref{eq:anatomical_offset_matrix} represents the link-based (anatomical) offsets with respect to the human partner $\mathcal{P}$ while being aware of itself $\mathcal{R}$. 

\paragraph{Joint Importance Weighting}
As certain parts of the body are more involved in the interactions, we propose joint importance weighting to prioritize them when retargeting.

We associate each joint $J_{i}^{\mathcal{H}}$ in the human $\mathcal{H}$ with a scalar weight using a softmax-normalized function~\cite{bahdanau2014neural} $w_{\mathrm{sem}, i}^{\mathcal{H}} = {\exp (\alpha_i)}/{\sum_{j=1}^{n} \exp (\alpha_j)}$,
where the cosine similarity $\alpha_i = (\mathbf{z}_{\mathrm{poses}} \cdot \mathbf{z}_{\mathrm{joint}})/(\lVert \mathbf{z}_{\mathrm{poses}} \rVert \lVert\mathbf{z}_{\mathrm{joint}} \rVert)$ with $\mathbf{z}_{\mathrm{poses}}$ and $\mathbf{z}_{\mathrm{joint}}$ respectively represent the embedding vectors of the poses throughout the interaction horizon and of the specific joint; both of which are obtained using MotionCLIP~\cite{tevet2022motionclip}. With the weight $w_{\mathrm{sem}}$ along the interaction horizon, the pairwise weight between $J_{i}^{\mathcal{H}}$ and $J_{j}^{\mathcal{P}}$ applied to all poses is $w_{\mathrm{sem}, ij}^{\mathcal{H} \rightarrow \mathcal{P}} = w_{\mathrm{sem}, i}^{\mathcal{H}} \, w_{\mathrm{sem}, j}^{\mathcal{P}}$, which forms the $n \times n$ weighting matrix:
\begin{equation}
    \mathbf{W}^{\mathcal{H} \rightarrow \mathcal{P}}_{\mathrm{sem}} = 
    \begin{bmatrix}
        w^{\mathcal{H} \rightarrow \mathcal{P}}_{\mathrm{sem}, 11} & w^{\mathcal{H} \rightarrow \mathcal{P}}_{\mathrm{sem}, 12} & \dots & w^{\mathcal{H} \rightarrow \mathcal{P}}_{\mathrm{sem}, 1n} \\
        \vdots & \vdots & \ddots & \vdots \\
        w^{\mathcal{H} \rightarrow \mathcal{P}}_{\mathrm{sem}, n1} & w^{\mathcal{H} \rightarrow \mathcal{P}}_{\mathrm{sem}, n2} & \dots & w^{\mathcal{H} \rightarrow \mathcal{P}}_{\mathrm{sem}, nn} \\
    \end{bmatrix}.
    \label{eq:pairwise_weighting_matrix}
\end{equation}

\paragraph{Joint-Relative Distance Weighting}
We additionally weigh the joint-relative distance between $J_{i, t}^{\mathcal{H}}$ and $J_{j, t}^{\mathcal{P}}$ at each discrete time $t$, modeled as a modified normal distribution: $w_{\mathrm{dis}, ij, t}^{\mathcal{H} \rightarrow \mathcal{P}} = \exp \left( -{\lVert r^{\mathcal{H} \rightarrow \mathcal{P}}_{ij, t}\rVert}^2/{2 \sigma^2} \right)$, where $\sigma^2$ denotes the distribution variance. Note that, since the joint-relative distances are non-negative while the normal distribution is infinitely two-sided, joints that are closer to each other are assigned a larger weight and vice versa. Similar to Eq.~\ref{eq:pairwise_weighting_matrix}, the time-dependent distance-weighted matrix is:
\begin{equation}
    \mathbf{W}^{\mathcal{H} \rightarrow \mathcal{P}}_{\mathrm{dis}, t} = 
    \begin{bmatrix}
        w^{\mathcal{H} \rightarrow \mathcal{P}}_{\mathrm{dis}, 11, t} & w^{\mathcal{H} \rightarrow \mathcal{P}}_{\mathrm{dis}, 12, t} & \dots & w^{\mathcal{H} \rightarrow \mathcal{P}}_{\mathrm{dis}, 1n, t} \\
        \vdots & \vdots & \ddots & \vdots \\
        w^{\mathcal{H} \rightarrow \mathcal{P}}_{\mathrm{dis}, n1, t} & w^{\mathcal{H} \rightarrow \mathcal{P}}_{\mathrm{dis}, n2, t} & \dots & w^{\mathcal{H} \rightarrow \mathcal{P}}_{\mathrm{dis}, nn, t} \\
    \end{bmatrix}.
    \label{eq:joint_direction_weighting_matrix}
\end{equation}

\subsubsection{Mimetic Embodiment Transfer Objective}
At each discrete timestep $t$, given $\mathbf{J}^{\mathcal{H} \rightarrow \mathcal{P}}_t$, $\mathbf{A}^{\mathcal{H} \rightarrow \mathcal{P}}_t$, $\mathbf{W}^{\mathcal{H} \rightarrow \mathcal{P}}_{\mathrm{dis}, t}$, and $\mathbf{W}^{\mathcal{H} \rightarrow \mathcal{P}}_{\mathrm{sem}}$, we formulate the mimetic embodiment transfer term, $E_{\text{interact}}(\mathcal{R}_t, \mathcal{P}_t, \mathcal{H}_{\mathrm{SMPL}, t})$, to align human-humanoid interaction with its human-human interaction as:
\begin{equation}
    \begin{aligned}
        & \left\| (\mathbf{W}^{\mathcal{H} \rightarrow \mathcal{P}}_{\mathrm{sem}} \, \mathbf{W}^{\mathcal{H} \rightarrow \mathcal{P}}_{\mathrm{dis}}) \odot (\mathbf{J}^{\mathcal{R} \rightarrow \mathcal{P}}_t - \mathbf{J}^{\mathcal{H} \rightarrow \mathcal{P}}_t) \right\|_{2}^{2} \\
        &+ \left\| (\mathbf{W}^{\mathcal{H} \rightarrow \mathcal{P}}_{\mathrm{sem}} \, \mathbf{W}^{\mathcal{H} \rightarrow \mathcal{P}}_{\mathrm{dis}}) \odot (\mathbf{A}^{\mathcal{R} \rightarrow \mathcal{P}}_t - \mathbf{A}^{\mathcal{H} \rightarrow \mathcal{P}}_t) \right\|_{2}^{2}, \\
    \end{aligned}
    \label{eq:met_term}
\end{equation}
where $\odot$ denotes element-wise multiplication. By incorporating Eq.~\ref{eq:met_term} with an enforced inverse kinematic term $E_{\mathrm{IK}}$ and the Tikhonov regularization term $E_{\mathrm{smooth}}$~\cite{wampler1986manipulator}, we therefore use the \texttt{mink} solver~\cite{minksolver} to solve for $\widehat{\mathcal{R}}_t$ in Eq.~\ref{eq:met_objective} at each discrete timestep $t$ along the interaction horizon of $T$:
\begin{equation}
    \begin{aligned}
        \widehat{\mathcal{R}}_t &= \arg \min_{\mathcal{R}_t} E_{\text{interact}}(\mathcal{R}_t, \mathcal{P}_t, \mathcal{H}_{\mathrm{SMPL}, t}) \\
        \qquad &+ E_{\text{IK}}(\mathcal{R}_t, \mathcal{H}_{\mathrm{SMPL}, t}) + E_{\text{smooth}}(\mathcal{R}_{t},\mathcal{R}_{t-1})
    \end{aligned}
    \label{eq:met_objective}
\end{equation}
to obtain $\{ \widehat{\mathcal{R}}_t \}_{t=0}^{T}$, preserving the interaction between the target human $\{ \mathcal{H}_{\mathrm{SMPL}, t} \}_{t=0}^{T}$ and its human partner $\{ \mathcal{P}_t \}_{t=0}^{T}$ along the interaction horizon.

\begin{figure*}[t]
    \centering
    \vspace{4pt}
    \includegraphics[width=1.00\linewidth]{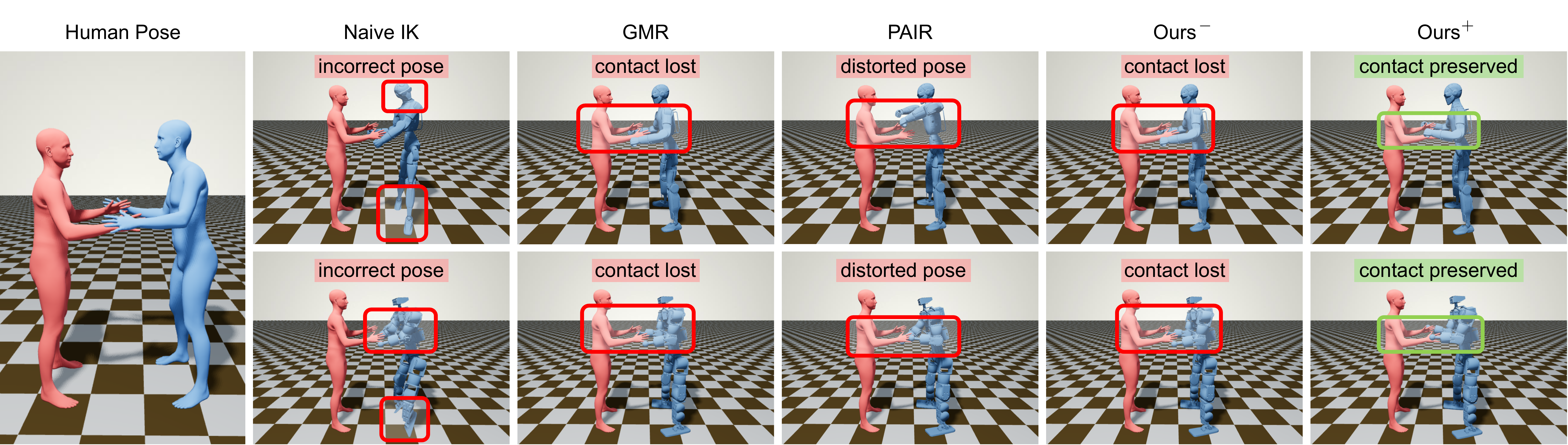}
    \vspace{-15pt}
    \caption{\textbf{Qualitative Comparisons of Retargeting Results:} Given the human-human interaction to retarget to different humanoids, naive IK~\cite{ProtoMotions} produces suboptimal leg and head poses on the Unitree R1 Edu and foot-ground penetration on the HighTorque Mini-hi. Meanwhile, GMR~\cite{joao2025gmr} and~\systemName$^{-}$(Ours) preserve robot poses but lose contact with the human partner, whereas PAIR~\cite{huang2026learning} introduces distorted poses on both humanoids.~\systemName$^{+}$(Ours) preserves embodiment-specific poses and maintains contact with the human partner.} 
    \vspace{-15pt}
    \label{fig:qualitative}
\end{figure*}

\vspace{-1pt}
\subsection{Post-Retargeting Interaction Optimization}
\label{sec:post_retargeting}
Yet, these newly retargeted poses face challenges in self-balancing and inter-body collisions. We mitigate these artifacts by optimizing post-retargeting interactions as we roll out their poses. For self-balancing, we follow the full-body compliant human-humanoid interaction mechanism~\cite{hyon2007full}. To prevent inter-body collisions, we maintain an inter-body minimum safe distance of $d_{\min, ij}$ between $J_{i, t}^{\mathcal{R}}$ and $J_{j, t}^{\mathcal{P}}$ at each given discrete timestep $t$. We apply this post-retargeting optimization throughout the human-humanoid interaction horizon for all human-humanoid interaction sequences. 

Lastly, we apply the entire~\systemName~framework to all human-human interaction sequences, yielding the final~\datasetName, which is used in our experiments.

\section{Experiment}
\label{sec:experiment}

We conduct intensive experiments to: (\textit{i}) evaluate the effectiveness of~\systemName; (\textit{ii}) demonstrate the utility of our~\datasetName~dataset for a generative human-humanoid interaction task; and (\textit{iii}) validate real-world deployment by using a trained policy on our~\datasetName~dataset and executing it on a real humanoid robot. 

\vspace{-4pt}
\subsection{Assessment of Extraction \& Retargeting in~\systemName}
First, we assess the performance of our~\systemName~in the human-to-humanoid motion transfer task. As no training is involved, all experiments in this section run on \textit{our full}~\datasetName~dataset.

\subsubsection{Physical Plausibility of Extracted Human-Human Interaction}
We first evaluate the physical plausibility of extracted human-human interactions (Eq.~\ref{eq:foot_consistent_error}) by computing their foot-slipping errors ($E_{\mathrm{foot}}$)~\cite{rempe2020contact}, ground penetration ($E_{\mathrm{ground}}$)~\cite{yuan2021simpoe}, and static stability margin (SSM)~\cite{silva2007historical}. As in~\cite{maoudj2026spinder}, we compute SSM as the signed distance between the projected center of mass (CoM) and the support polygon. 

\begin{table}[h]
    \centering
    \setlength{\tabcolsep}{5pt}
    \caption{\textbf{Effects of Foot-Ground Consistency Fine-tuning:} Quantitative comparisons between SMPL-based fitting and itself with our proposed foot-ground consistency fine-tuning (Eq.~\eqref{eq:foot_consistent_error}) in terms of $E_{\mathrm{foot}}$, $E_{\mathrm{ground}}$, and SSM on~\datasetName.}
    \vspace{-2pt}
    \resizebox{0.90\linewidth}{!}{
        \begin{tabular}{l|ccc}
            \toprule
            \diagbox{Method}{Metric} & \makecell{$E_{\mathrm{foot}}$\\($\downarrow$)} & \makecell{$E_{\mathrm{ground}}$\\($\downarrow$)} & \makecell{SSM\\($\uparrow$)} \\
            \midrule \midrule
            SMPL Fitting~\cite{pavlakos2019expressive} & \underline{2.6125} & \underline{36.3547} & \underline{-1.9336} \\
            \rowcolor[HTML]{EFEFEF} \makecell[l]{SMPL Fitting~\cite{pavlakos2019expressive}  with Foot-\\Ground Consistency Fine-tuning} & \textbf{0.6765} & \textbf{0.9282} & \textbf{2.0402} \\
            \bottomrule
        \end{tabular}
    }
    \vspace{-13pt}
    \label{tab:human_foot_finetune}
\end{table}

As depicted in Table~\ref{tab:human_foot_finetune}, our method achieves lower error across the defined metrics compared to the traditional SMPL-based fitting~\cite{pavlakos2019expressive}, notably with only $E_{\mathrm{foot}}$ of $0.6765$~\unit{cm}/\unit{frame}, $E_{\mathrm{ground}}$ of $0.9282$~\unit{cm}/\unit{frame}, and SSM of $2.0402$~\unit{cm}/\unit{frame} lying inside the polygon, highlighting the importance of foot-ground consistency fine-tuning.

\subsubsection{Mimetic Embodiment Transfer}
We evaluate the performance of our mimetic embodiment transfer against naive methods (\textit{i.e.}, inverse-kinematics~\cite{ProtoMotions} with uniform scaling) and optimization-based fitting methods (\textit{i.e.}, GMR~\cite{joao2025gmr} and PAIR~\cite{huang2026learning}) on the $23$-DoF HighTorque Mini-hi and the $26$-DoF Unitree R1 Edu. In this experiment, we report MPJPE in~\unit{cm} as the joint position error, IRD~\unit{cm} as the mean absolute pairwise-distance error, CF$1$ as the F$1$ score for contact preservation at a distance threshold of $d=20$~\unit{cm}. 

\begin{table}[h]
    \centering
    \setlength{\tabcolsep}{2pt}
    \vspace{-6pt}
    \caption{\textbf{Retargeting Performance on Different Humanoids:} Quantitative evaluations in terms of MPJPE, IRD, and CF$1$ on~\datasetName~between our methods against naive IK~\cite{ProtoMotions}, GMR~\cite{joao2025gmr}, and PAIR~\cite{huang2026learning} on the Unitree R1 Edu and the HighTorque Mini-hi.~\systemName$^{-}$ and~\systemName$^{+}$ denote for without and with the weighting matrices (Eq.~\ref{eq:pairwise_weighting_matrix} and Eq.~\ref{eq:joint_direction_weighting_matrix}).}
    \vspace{-2pt}
    \resizebox{1.00\linewidth}{!}{
        \begin{tabular}{l ccc ccc}
            \toprule
            \multirow{2}{*}{Method} & \multicolumn{3}{c}{Unitree R1 Edu} & \multicolumn{3}{c}{HighTorque Mini-hi} \\
            \cmidrule(lr){2-4} \cmidrule(lr){5-7} & \makecell{MPJPE ($\downarrow$)} & \makecell{IRD ($\downarrow$)} & \makecell{CF1 ($\uparrow$)} & \makecell{MPJPE ($\downarrow$)} & \makecell{IRD ($\downarrow$)} & \makecell{CF1 ($\uparrow$)} \\
            \midrule \midrule
            Naive IK~\cite{ProtoMotions} & 37.85 & 16.82 & 13.84 & 41.31 & 20.67  & 11.42 \\
            GMR~\cite{joao2025gmr} & 11.91 & 5.73 & 69.12 & \underline{10.49}  & \underline{5.30} &  61.97\\
            PAIR~\cite{huang2026learning} & \underline{9.37} & 4.16 & 74.54 & 12.79 & 6.24 & 59.41 \\
            \midrule
            \systemName$^{-}$ & 10.13 & \underline{4.08} & \underline{80.90} & 11.63 & 6.14 & \underline{65.29} \\
            \rowcolor[HTML]{EFEFEF} \textbf{\systemName$^{+}$} & \textbf{6.31} & \textbf{2.84} & \textbf{81.97} & \textbf{7.58}  & \textbf{3.67} & \textbf{78.15} \\
            \bottomrule
        \end{tabular}
    }
    \vspace{-10pt}
    \label{tab:ik_baseline}
\end{table}

In Table~\ref{tab:ik_baseline}, our method with semantic and distance-based weighting (\systemName$^{+}$) achieves the best MPJPE, IRD, and CF$1$ results across both the Unitree R1 Edu and the HighTorque Mini-hi. Meanwhile, GMR~\cite{joao2025gmr}, PAIR~\cite{huang2026learning}, and our method without a weighting mechanism (\systemName$^{-}$) result in slight inconsistencies. The qualitative results in Fig.~\ref{fig:qualitative} confirm the overall effectiveness of our weighting strategy for preserving high-fidelity, embodiment-specific human-to-humanoid interaction retargeting.

\begin{figure}[ht!]
    \centering
    \vspace{4pt}
    \setlength{\tabcolsep}{4pt}
    \includegraphics[width=1.00\linewidth]{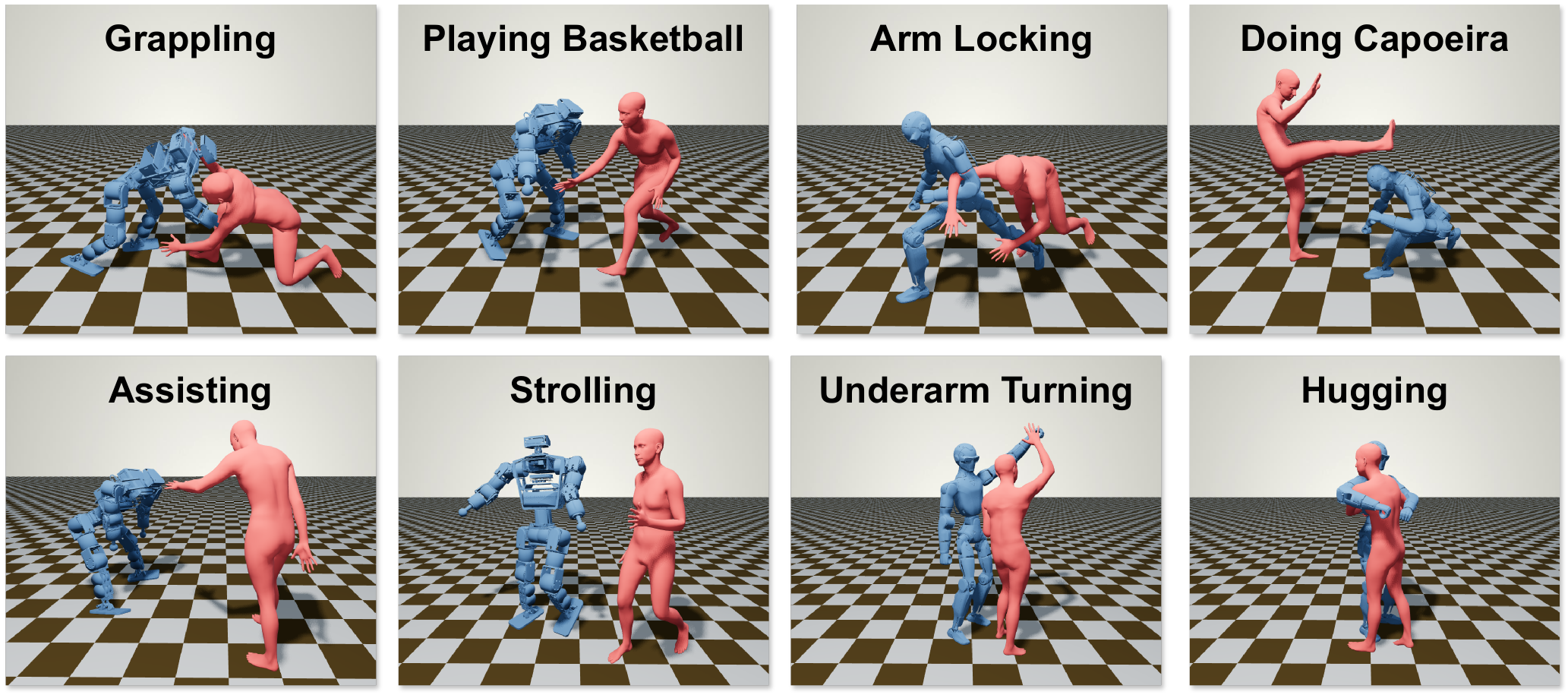}
    \vspace{-15pt}
    \caption{\textbf{Qualitative Samples of Retargeted Human-Humanoid Interactions:} By leveraging internet-scale human-human interaction videos,~\datasetName~exhibits numerous human-humanoid interactions. More samples are available in our supplementary video.}
    \vspace{-18pt}
    \label{fig:result_Vis}
\end{figure}

\subsubsection{Diversity of Retargeted Human-Humanoid Motion}
We also measure the motion diversity via the diversity score proposed in~\cite{guo2022generating}. As shown in Table~\ref{tab:diversity}, our dataset achieves higher diversity scores of $6.2427$ and $6.2285$ for Unitree R1 Edu and HighTorque Mini-hi, respectively, compared to AMASS~\cite{mahmood2019amass}'s diversity scores of $5.4834$ and $5.4859$. These results indicate greater overall motion variability, thanks to the collection of in-the-wild human-human interaction videos. The diversity of interactions can be seen in our video.
In addition, we show in Fig.~\ref{fig:result_Vis} the eight most representative human-humanoid interaction motions in~\datasetName.

\begin{table}[h]
    \centering
    \vspace{-6pt}
    \caption{\textbf{Motion Diversity of Datasets on Different Humanoids:} Diversity scores~\cite{guo2022generating} of human-humanoid interactions on~\datasetName~compared to Human3.6M~\cite{ionescu2013human3}, AMASS~\cite{mahmood2019amass}, and 3DPW~\cite{von20183dpw} on the Unitree R1 Edu and the HighTorque Mini-hi.}
    \vspace{-2pt}
    \resizebox{0.82\linewidth}{!}{
        \begin{tabular}{l cc}
            \toprule
            \multirow{2}{*}{Method} & Unitree R1 Edu & HighTorque Mini-hi \\
            \cmidrule(lr){2-2} \cmidrule(lr){3-3} & \makecell{Diversity ($\uparrow$)} & \makecell{Diversity ($\uparrow$)} \\
            \midrule \midrule
            Human3.6M~\cite{ionescu2013human3} & 3.8251 & 3.7963 \\
            AMASS~\cite{mahmood2019amass} & \underline{5.4834} & \underline{5.4859} \\
            3DPW~\cite{von20183dpw} & 4.0173 & 3.9130 \\
            \midrule
            \rowcolor[HTML]{EFEFEF} \textbf{\datasetName} & \textbf{6.2427} & \textbf{6.2285} \\
            \bottomrule
        \end{tabular}
    }
    \vspace{-13pt}
    \label{tab:diversity}
\end{table}

\subsection{Generative Human-Humanoid Interaction Benchmarking}
Based on the concept of mimesis, we verify whether our \datasetName~dataset can serve as a benchmark for generative human-humanoid interaction tasks. In this experiment, we split our~\datasetName~dataset into train and test sets in an $8:2$ ratio to train and test existing baselines. Given the human's motion sequence, the generative human-humanoid interaction task aims to synthesize a responsive 3D motion for the humanoid to synchronize with the partner's motion while maintaining physical and semantic coordination, including handling occlusions, rotations, and physical contacts.

We evaluate on five architectures: ReMos~\cite{ghosh2024remos}, InterFormer~\cite{chopin2023interaction}, ComMDM~\cite{shafir2023human}, InterGen~\cite{liang2024intergen}, and Duolando~\cite{siyaoduolando}, adapted to predict follower motion conditioned on partner's 3D poses. For Duolando, RL fine-tuning further reduces artifacts such as foot slipping and penetration. 
Following \cite{siyaoduolando}, we evaluate the motion kinematic quality through the kinematic Fréchet Inception Distance \cite{heusel2017gans} ($\mathrm{FID}_{k}$) from joint trajectories and interaction quality through cross-distance FID ($\mathrm{FID}_{cd}$) between pairwise joints and frequency of contact consistency ($f_{\text{contact}}$) in terms of percentages of contact frames versus non-contact frames.

\begin{figure}[t]
    \vspace{4pt}
    \centering
    \scriptsize
    \includegraphics[width=1.00\linewidth]{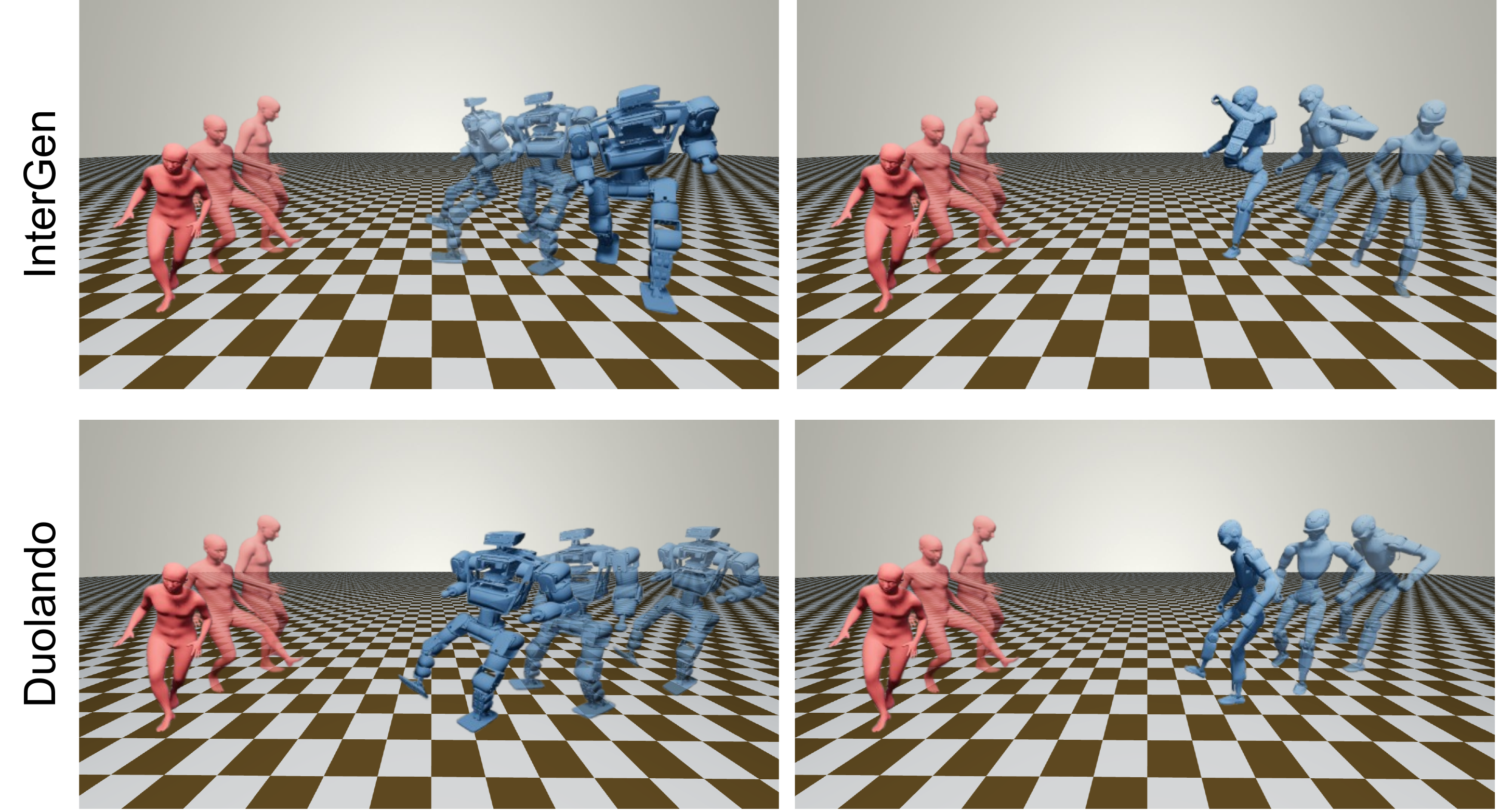}
    \vspace{-10pt}
    \caption{\textbf{Qualitative Samples of Generative Human-Humanoid Interaction:} Qualitative comparison of humanoid-human interaction generation on the test set of~\datasetName. The context of the input is ``\textit{A man is kicking a ball}''; the humanoid is expected to move to receive the ball as it is passed. Duolando~\cite{siyaoduolando} achieves a more natural and reactive response to follower motions, while InterGen~\cite{liang2024intergen} either mimics the partner's motion or responds with an unsuitable motion. More qualitative results of other baselines are provided in our supplementary video.}
    \vspace{-18pt}
    \label{fig:hhi_qualitative}
\end{figure}

Table~\ref{tab:hhi_results} shows that Duolando achieves the best overall performance, with FID$_k$ at $24.96$ and $27.24$, FID$_{cd}$ at $9.07$ and $11.53$, together with the $f_{\text{contact}}$ of $51.75\%$ and $46.77\%$. ReMos~\cite{ghosh2024remos} exhibits improved contact frequency at $46.79\%$ and $47.56\%$. Meanwhile, InterGen~\cite{liang2024intergen} achieves an FID$_{cd}$ of $10.86$ under HighTorque Mini-hi embodiment. Fig.~\ref{fig:hhi_qualitative} presents a challenging sports sequence with a human partner. Duolando~\cite{siyaoduolando} produces more appropriate and synchronized responses, while InterGen~\cite{liang2024intergen} generates coherent motion but often responds inappropriately, either closely mimicking the human partner or producing unnatural motions.

\begin{table}[!ht]
    \centering
    \setlength{\tabcolsep}{2pt}
    \vspace{-6pt}
    \caption{\textbf{Generating Humanoid Motion from Human Motion Results.} Quantitative results on humanoid-human interaction between ComMDM~\cite{shafir2023human}, InterFormer~\cite{chopin2023interaction}, InterGen~\cite{liang2024intergen}, ReMos~\cite{ghosh2024remos}, and Duolando~\cite{siyaoduolando}. All experiments run on the test set of our~\datasetName~dataset.}
    \vspace{-2pt}
    \resizebox{1.00\linewidth}{!}{
        \begin{tabular}{l ccc ccc}
            \toprule
            \multirow{2}{*}{Method} & \multicolumn{3}{c}{Unitree R1 Edu} & \multicolumn{3}{c}{HighTorque Mini-hi} \\
            \cmidrule(lr){2-4} \cmidrule(lr){5-7} & \makecell{FID$_k$} ($\downarrow$) & \makecell{FID$_{cd}$} ($\downarrow$) & \makecell{$f_{\text{contact}}$} ($\uparrow$) & \makecell{FID$_k$} ($\downarrow$) & \makecell{FID$_{cd}$} ($\downarrow$) & \makecell{$f_{\text{contact}}$} ($\uparrow$) \\
            \midrule \midrule
            ComMDM~\cite{shafir2023human} & 36.07 & 15.93 & 31.09\% & 38.33 & 18.30 & 29.07\% \\
            InterFormer~\cite{chopin2023interaction} & 30.16 & 14.66 & 42.97\% & \underline{28.64} & 13.78 & 37.49\% \\
            InterGen~\cite{liang2024intergen} & 34.62 & 17.72 & \underline{48.91\%} & 30.98 & \textbf{10.86} & 39.37\% \\
            ReMos~\cite{ghosh2024remos} & \underline{25.49} & \underline{11.66} & 46.79\% & 31.51 & 12.94 & \textbf{47.56\%} \\
            Duolando~\cite{siyaoduolando} & \textbf{24.96} & \textbf{9.07} & \textbf{51.75\%} & \textbf{27.24} & \underline{11.53} & \underline{46.77\%} \\
            \bottomrule
        \end{tabular}
    }
    \vspace{-13pt}
    \label{tab:hhi_results}
\end{table}

\subsection{Self-Balancing Evaluation \& Real-Robot Deployment}
Finally, we examine the self-balancing capability of the Unitree R1 Edu and the HighTorque Mini-hi in simulation and physically demonstrate the interaction tasks on the real Unitree R1 Edu humanoid with a human partner.

\begin{figure*}[t]
    \centering
    \vspace{4pt}
    \includegraphics[width=0.98\linewidth]{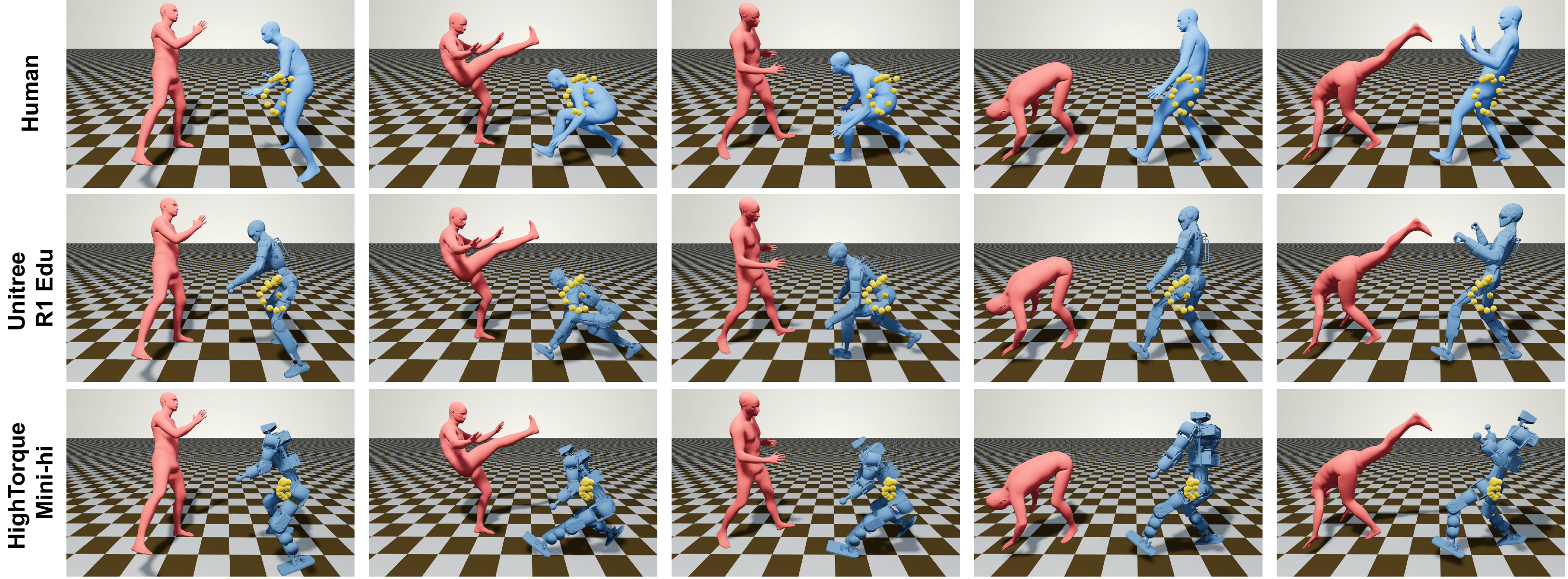}
    \vspace{-2pt}
    \caption{\textbf{Visualization of CoM Trajectories on Different Humanoid Embodiments:} Qualitative comparison of the original human motion (top) and the retargeted motions to Unitree R1 Edu (middle) and HighTorque Mini-hi (bottom) during the kicking interaction. The yellow markers visualize CoM trajectories of humans and humanoids, showing their geometric similarities to the human's CoM trajectories.} 
    \vspace{-13pt}
    \label{fig:com_qualitative}
\end{figure*}

\subsubsection{Self-Balancing Evaluation}
We evaluate the balancing ability of the retargeted motions for both humanoid embodiments, considering their CoM heights relative to the ground's tangential plane, and we compare these profiles with those of the retargeted human along the interaction horizon.

On the test set of~\datasetName~in simulation, both humanoids, Unitree R1 Edu and HighTorque Mini-hi, maintain CoM-height differences of $2.9573$ \unit{cm} and $3.2021$ \unit{cm}, respectively, while preserving interaction-critical relationships with the human partner. 
Both the Unitree R1 Edu (middle) and HighTorque Mini-hi (bottom) in Fig.~\ref{fig:com_qualitative} are capable of reproducing
the crouching motions while maintaining the retargeted human's CoM trajectories and remaining consistent with their respective morphologies, as shown in Fig.~\ref{fig:com_qualitative} (top). These results indicate that~\systemName~can adapt human poses to each humanoid while maintaining self-balancing capabilities, which are well-suited for real-robot deployment.

\begin{figure}[t]
    \centering
    \centering
    \vspace{4pt}
\includegraphics[width=0.95\linewidth]{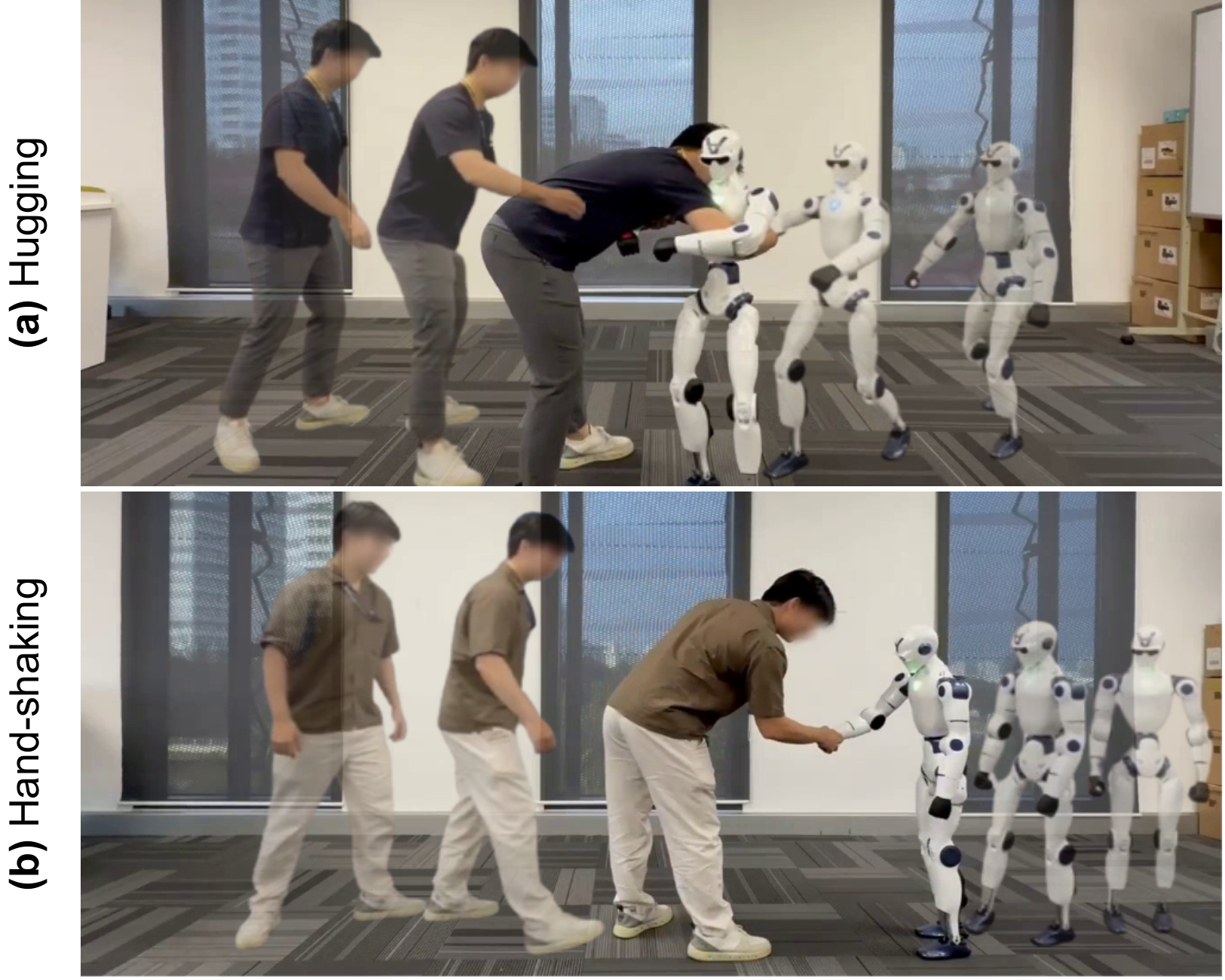}
    \vspace{-3pt}
    \caption{\textbf{Real-Robot Deployment on the Unitree R1 Edu:} Two deployed human-humanoid interaction sequences on the Unitree R1 Edu: \textbf{(a)} hugging and \textbf{(b)} hand-shaking with overlaid snapshots depict the temporal evolution of the interacting motion on both the humanoid and its human partner. The interacting sequences are safe, and the humanoid does not cause harm in either scenario.}
    \vspace{-20pt}
    \label{fig:realDemo}
\end{figure}

\subsubsection{Real-Robot Deployment} To evaluate real-world executability, we deploy the retargeted motions on the $26$-DoF Unitree R1 Edu. 
We use the sequences from~\datasetName~to optimize and map to the robot's joints, then check against the robot's motion constraints before execution. In particular, we employ the Unitree R1 Edu motion-tracking controller in \texttt{mjlab}~\cite{zakka2026mjlab}, thus optimizing a $26$-action policy in $4096$ parallel environments for approximately $30,000$ iterations on an NVIDIA RTX 3090 GPU.

The training stage samples initial motion phases, perturbs the robot's initial pose and velocity, and randomizes physical parameters, including foot friction and joint dynamics. Following DeepMimic~\cite{peng2018deepmimic} and BeyondMimic~\cite{liao2025beyondmimic}, we condition the actor on this command and proprioception, rewarding the simulated robot for matching the reference. At each simulated step, a motion command samples joint positions and velocities to form the policy command, while anchor and whole-body states supply the corresponding tracking targets. The trained actor is then exported to ONNX and executed in the C++ mimic controller with ONNX Runtime. At $50$~Hz, the controller reads the $26$ motor states and IMU from Unitree low-state messages, reconstructs the training observation order, and finally evaluates the policy. For details, the observations contain reference joint positions and velocities, torso orientation error, IMU angular velocity, and the previous motion. The torso orientation is reconstructed from the IMU quaternion and waist joints without global position estimation. 
A fixed mapping converts the common $26$-joint policy order to the sparse Unitree SDK motor IDs used on the robot platform. Fig.~\ref{fig:realDemo} shows that our framework preserves the characteristic spatial and contact relationships in retargeted human-human interactions. The demonstration video is available in our supplementary materials.


\vspace{-2pt}
\subsection{Limitations \& Future Work}
Nevertheless,~\systemName~remains limited by substantial morphological differences between humans and humanoids, which can make certain motions kinematically infeasible. For example, in simulation, we found that the HighTorque Mini-hi's broad shoulders and short arms prevent it from hugging a human partner as the Unitree R1 Edu can. Furthermore, our current real-robot deployment primarily relies on predefined motion sequences and does not yet explicitly account for human motion. An interesting direction for future work is therefore to develop methods that adapt humanoid motion in response to observed human behavior, leveraging motion data from our~\datasetName~dataset.

\section{Conclusion}
\label{sec:conclusion}
In this paper, we have presented~\systemName~for mimetic embodiment transfer from in-the-wild human-human interaction videos. Our proposed method includes a data preparation pipeline that extracts physically plausible human-human interaction poses from in-the-wild videos, followed by a mimetic embodiment transfer that translates one of the observed humans into a humanoid. Our~\datasetName~dataset demonstrates higher quality in avoiding ground penetration and foot slipping, while~\systemName~allows human poses to be retargeted to different humanoids. Benchmarking on~\datasetName~across motion generation methods and deployments on the Unitree R1 Edu humanoid showed that our work enables physical deployment and can serve as a benchmark for future human-humanoid interaction work.

\bibliographystyle{IEEEtran}
\bibliography{reference}

@article{jiang2024harmon,
	title        = {Harmon: Whole-body motion generation of humanoid robots from language descriptions},
	author       = {Jiang, Zhenyu and others},
	year         = 2024,
	journal      = {arXiv}
}

@inproceedings{he2024learning,
	title        = {Learning human-to-humanoid real-time whole-body teleoperation},
	author       = {He, Tairan and others},
	year         = 2024,
	booktitle    = {IROS}
}

@article{he2024omnih2o,
	title        = {Omnih2o: Universal and dexterous human-to-humanoid whole-body teleoperation and learning},
	author       = {He, Tairan and others},
	year         = 2024,
	journal      = {arXiv}
}

@inproceedings{pavlakos2019expressive,
	title        = {Expressive body capture: 3d hands, face, and body from a single image},
	author       = {Pavlakos, Georgios and others},
	year         = 2019,
	booktitle    = {CVPR}
}

@inproceedings{sun2022_dancetrack,
	title        = {DanceTrack: Multi-Object Tracking in Uniform Appearance and Diverse Motion},
	author       = {Peng Sun and others},
	year         = 2022,
	booktitle    = {CVPR}
}

@inproceedings{kajita2003biped,
	title        = {Biped Walking Pattern Generation by using Preview Control of Zero-Moment Point},
	author       = {Kajita, Shuuji and others},
	year         = 2003,
	booktitle    = {ICRA}
}

@article{ionescu2013human3,
	title        = {Human3.6M: Large Scale Datasets and Predictive Methods for 3D Human Sensing in Natural Environments},
	author       = {Ionescu, Catalin and others},
	year         = 2014,
	journal      = {TPAMI}
}

@misc{cmu_mocap,
	title        = {CMU Graphics Lab Motion Capture Database},
	author       = {CMU},
	year         = 2003,
}

@inproceedings{mahmood2019amass,
	title        = {AMASS: Archive of Motion Capture as Surface Shapes},
	author       = {Mahmood, Naureen and Ghorbani, Nima and Troje, Nikolaus F. and Pons-Moll, Gerard and Black, Michael J.},
	year         = 2019,
	booktitle    = {ICCV}
}

@inproceedings{loper2015smpl,
	title        = {SMPL: A Skinned Multi-Person Linear Model},
	author       = {Loper, Matthew and Mahmood, Naureen and Romero, Javier and Pons-Moll, Gerard and Black, Michael J.},
	year         = 2015,
	booktitle    = {TOG}
}

@inproceedings{fieraru2021three,
	title        = {Three-Dimensional Reconstruction of Human Interactions},
	author       = {Fieraru, Mihai and others},
	year         = 2021,
	booktitle    = {CVPR}
}

@inproceedings{von20183dpw,
    title = {Recovering Accurate 3D Human Pose in The Wild Using IMUs and a Moving Camera},
    author = {von Marcard, Timo and Henschel, Roberto and Black, Michael and Rosenhahn, Bodo and Pons-Moll, Gerard},
    booktitle = {ECCV},
    year = {2018},
    month = {sep}
}

@inproceedings{mandery2015kit,
	title        = {The {KIT} Whole-Body Human Motion Database},
	author       = {Mandery, Christian and Terlemez, Omer and Do, Martin and Vahrenkamp, Nikolaus and Asfour, Tamim},
	year         = 2015,
	booktitle    = {ICAR}
}

@article{aristidou2011fabrik,
	title        = {FABRIK: A Fast, Iterative Solver for the Inverse Kinematics Problem},
	author       = {Aristidou, Andreas and Lasenby, Joan},
	year         = 2011,
	journal      = {Graphical Models}
}

@inproceedings{peng2018deepmimic,
	title        = {DeepMimic: Example-Guided Deep Reinforcement Learning of Physics-Based Character Skills},
	author       = {Peng, Xue Bin and others},
	year         = 2018,
	booktitle    = {TOG}
}

@inproceedings{siyaoduolando,
	title        = {Duolando: Follower GPT with Off-Policy Reinforcement Learning for Dance Accompaniment},
	author       = {Siyao, Li and others},
	year         = 2024,
	booktitle    = {ICLR}
}

@article{khirodkar2024harmony4d,
	title        = {Harmony4d: A video dataset for in-the-wild close human interactions},
	author       = {Khirodkar, Rawal and Song, Jyun-Ting and Cao, Jinkun and Luo, Zhengyi and Kitani, Kris},
	year         = 2024,
	journal      = {NeurIPS}
}

@inproceedings{ghosh2024remos,
	title        = {Remos: 3d motion-conditioned reaction synthesis for two-person interactions},
	author       = {Ghosh, Anindita and others},
	year         = 2024,
	booktitle    = {ECCV}
}

@article{chopin2023interaction,
	title        = {Interaction transformer for human reaction generation},
	author       = {Chopin, Baptiste and others},
	year         = 2023,
	journal      = {IEEE Transactions on Multimedia}
}

@article{shafir2023human,
	title        = {Human motion diffusion as a generative prior},
	author       = {Shafir, Yonatan and Tevet, Guy and Kapon, Roy and Bermano, Amit H},
	year         = 2023,
	journal      = {arXiv}
}

@article{liang2024intergen,
	title        = {Intergen: Diffusion-based multi-human motion generation under complex interactions},
	author       = {Liang, Han and Zhang, Wenqian and Li, Wenxuan and Yu, Jingyi and Xu, Lan},
	year         = 2024,
	journal      = {International Journal of Computer Vision}
}

@inproceedings{trumble2017total,
  title={Total capture: 3d human pose estimation fusing video and inertial sensors},
  author={Trumble, Matthew and others},
  booktitle={BMVC},
  year={2017}
}

@inproceedings{joo2015panoptic,
  title={Panoptic studio: A massively multiview system for social motion capture},
  author={Joo, Hanbyul and others},
  booktitle={ICCV},
  year={2015}
}

@article{luo2022embodied,
  title={Embodied scene-aware human pose estimation},
  author={Luo, Zhengyi and Iwase, Shun and Yuan, Ye and Kitani, Kris},
  journal={NeurIPS},
  year={2022}
}

@inproceedings{cheng2019occlusion,
  title={Occlusion-aware networks for 3d human pose estimation in video},
  author={Cheng, Yu and Yang, Bo and Wang, Bo and Yan, Wending and Tan, Robby T},
  booktitle={ICCV},
  year={2019}
}

@article{bandura1961transmission,
  title={Transmission of aggression through imitation of aggressive models.},
  author={Bandura, Albert and Ross, Dorothea and Ross, Sheila A},
  journal={The Journal of Abnormal and Social Psychology},
  volume={63},
  number={3},
  pages={575},
  year={1961},
  publisher={American Psychological Association}
}

@inproceedings{nakamura2007mimetic,
  title={Mimetic communication theory for humanoid robots interacting with humans},
  author={Nakamura, Yoshihiko and others},
  booktitle={Robotics Research: Results of the 12th International Symposium ISRR},
  pages={128--139},
  year={2007},
  organization={Springer}
}

@article{inamura2004embodied,
  title={Embodied symbol emergence based on mimesis theory},
  author={Inamura, Tetsunari and Toshima, Iwaki and Tanie, Hiroaki and Nakamura, Yoshihiko},
  journal={The International Journal of Robotics Research},
  volume={23},
  number={4-5},
  pages={363--377},
  year={2004},
  publisher={SAGE Publications}
}

@article{nguyen2025td,
  title={TD-GRPC: Temporal difference learning with group relative policy constraint for humanoid locomotion},
  author={Nguyen, Khang and Nguyen, Khai and Le, An T and Peters, Jan and Huber, Manfred and Vien, Ngo Anh and Vu, Minh Nhat},
  journal={arXiv preprint arXiv:2505.13549},
  year={2025}
}

@article{nguyen2025doublyaware,
  title={DoublyAware: Dual planning and policy awareness for temporal difference learning in humanoid locomotion},
  author={Nguyen, Khang and Le, An T and Peters, Jan and Vu, Minh Nhat},
  journal={IEEE Robotics and Automation Letters},
  volume={11},
  number={2},
  pages={2162--2169},
  year={2025},
  publisher={IEEE}
}

@article{hyon2007full,
  title={Full-body compliant human--humanoid interaction: balancing in the presence of unknown external forces},
  author={Hyon, SangHo and others},
  journal={IEEE Transactions on Robotics},
  volume={23},
  number={5},
  pages={884--898},
  year={2007},
  publisher={IEEE}
}

@inproceedings{yin2024whac,
  title={Whac: World-grounded humans and cameras},
  author={Yin, Wanqi and others},
  booktitle={ECCV},
  year={2024},
}

@inproceedings{sampieri2022poseCHICO,
  title={Pose forecasting in industrial human-robot collaboration},
  author={Sampieri, Alessio and others},
  booktitle={ECCV},
  year={2022},
}

@article{bu2025agibot,
  title={Agibot world colosseo: A large-scale manipulation platform for scalable and intelligent embodied systems},
  author={Bu, Qingwen and Cai, Jisong and Chen, Li and Cui, Xiuqi and Ding, Yan and Feng, Siyuan and Gao, Shenyuan and He, Xindong and Hu, Xuan and Huang, Xu and others},
  journal={ICRA},
  year={2025}
}

@inproceedings{li2025clone,
  title={Clone: Closed-loop whole-body humanoid teleoperation for long-horizon tasks},
  author={Li, Yixuan and Lin, Yutang and Cui, Jieming and Liu, Tengyu and Liang, Wei and Zhu, Yixin and Huang, Siyuan},
  booktitle={9th Annual Conference on Robot Learning},
  year={2025}
}

@article{joao2025gmr,
  title={Retargeting Matters: General Motion Retargeting for Humanoid Motion Tracking},
  author= {Joao Pedro Araujo and others},
  year= {2025},
  journal= {arXiv preprint arXiv:2510.02252}
}

@article{wang2026magic,
  title={MAGIC: Mixed Reality-Enhanced Active Constraint Generation and Impedance Control for Human-Robot Co-Manipulation},
  author={Wang, Tengyue and others},
  journal={IEEE Transactions on Automation Science and Engineering},
  year={2026},
  publisher={IEEE}
}

@article{jiang2025kaiwu,
  title={Kaiwu: A multimodal manipulation dataset and framework for robot learning and human-robot interaction},
  author={Jiang, Shuo and others},
  journal={IEEE Robotics and Automation Letters},
  year={2025},
  publisher={IEEE}
}

@inproceedings{bogo2016keep,
  title={Keep it SMPL: Automatic estimation of 3D human pose and shape from a single image},
  author={Bogo, Federica and others},
  booktitle={ECCV},
  pages={561--578},
  year={2016},
  organization={Springer}
}

@inproceedings{shin2024wham,
  title={Wham: Reconstructing world-grounded humans with accurate 3d motion},
  author={Shin, Soyong and others},
  booktitle={CVPR},
  pages={2070--2080},
  year={2024},
  organization={IEEE}
}

@article{huang2026human2humanoid,
  title={Human2Humanoid: Physics-Aware Cross-Morphology Motion Retargeting for Humanoid Robots},
  author={Huang, Tianchen and others},
  journal={arXiv preprint arXiv:2606.03476},
  year={2026}
}

@misc{ProtoMotions,
  title = {ProtoMotions3: An Open-source Framework for Humanoid Simulation and Control},
  author = {Tessler*, Chen and Jiang*, Yifeng and others},
  year = {2025},
  publisher = {GitHub},
  journal = {GitHub repository},
}

@article{huang2026learning,
  title={Learning whole-body human-humanoid interaction from human-human demonstrations},
  author={Huang, Wei-Jin and others},
  journal={arXiv preprint arXiv:2601.09518},
  year={2026}
}

@inproceedings{choi2021self,
  title={Self-supervised motion retargeting with safety guarantee},
  author={Choi, Sungjoon and others},
  booktitle={ICRA},
  pages={8097--8103},
  year={2021},
  organization={IEEE}
}

@inproceedings{cheynel2025reconform,
  title={ReConForM: Real-time Contact-aware Motion Retargeting for more Diverse Character Morphologies},
  author={Cheynel, Th{\'e}o and others},
  booktitle={Computer Graphics Forum},
  volume={44},
  number={2},
  pages={e70028},
  year={2025},
  organization={Wiley Online Library}
}

@article{jang2024geometry,
  title={Geometry-aware retargeting for two-skinned characters interaction},
  author={Jang, Inseo and others},
  journal={TOG},
  volume={43},
  number={6},
  pages={1--17},
  year={2024},
  publisher={ACM New York, NY, USA}
}

@inproceedings{tevet2022motionclip,
  title={Motionclip: Exposing human motion generation to clip space},
  author={Tevet, Guy and others},
  booktitle={ECCV},
  pages={358--374},
  year={2022},
  organization={Springer}
}

@article{liao2025beyondmimic,
  title={Beyondmimic: From motion tracking to versatile humanoid control via guided diffusion},
  author={Liao, Qiayuan and Truong, Takara E and Huang, Xiaoyu and Gao, Yuman and Tevet, Guy and Sreenath, Koushil and Liu, C Karen},
  journal={arXiv preprint arXiv:2508.08241},
  year={2025}
}

@inproceedings{yuan2021simpoe,
  title={Simpoe: Simulated character control for 3d human pose estimation},
  author={Yuan, Ye and others},
  booktitle={CVPR},
  pages={7155--7165},
  year={2021},
  organization={IEEE}
}

@article{silva2007historical,
  title={A historical perspective of legged robots},
  author={Silva, Manuel F and others},
  journal={Journal of Vibration and Control},
  volume={13},
  number={9-10},
  pages={1447--1486},
  year={2007},
  publisher={Sage Publications Sage UK: London, England}
}

@article{maoudj2026spinder,
  title={SPINDER: an open-source 18-DoF hexapod robot with hierarchical central pattern generator control and analytic inverse kinematics},
  author={Maoudj, Abderraouf and others},
  journal={Scientific Reports},
  year={2026},
  publisher={Nature Publishing Group}
}

@inproceedings{rempe2020contact,
  title={Contact and human dynamics from monocular video},
  author={Rempe, Davis and others},
  booktitle={ECCV},
  pages={71--87},
  year={2020}
}

@inproceedings{guo2022generating,
  title={Generating diverse and natural 3d human motions from text},
  author={Guo, Chuan and others},
  booktitle={CVPR},
  pages={5142--5151},
  year={2022},
  organization={IEEE}
}

@article{heusel2017gans,
  title={Gans trained by a two time-scale update rule converge to a local nash equilibrium},
  author={Heusel, Martin and others},
  journal={NeurIPS},
  volume={30},
  year={2017}
}

@article{salazar2026xmopipe,
  title={XmoPipe: A Pipeline for Large-Scale In-the-Wild Human Motion Dataset Construction},
  author={Salazar, Nathan and others},
  journal={Computer Animation and Virtual Worlds},
  volume={37},
  number={4},
  pages={e70156},
  year={2026},
  publisher={Wiley Online Library}
}

@inproceedings{tang2024humanmimic,
  title={Humanmimic: Learning natural locomotion and transitions for humanoid robot via wasserstein adversarial imitation},
  author={Tang, Annan and others},
  booktitle={ICRA},
  pages={13107--13114},
  year={2024},
  organization={IEEE}
}

@article{bahdanau2014neural,
  title={Neural machine translation by jointly learning to align and translate},
  author={Bahdanau, Dzmitry and others},
  journal={arXiv preprint arXiv:1409.0473},
  year={2014}
}

@inproceedings{luo2023perpetual,
  title={Perpetual humanoid control for real-time simulated avatars},
  author={Luo, Zhengyi and others},
  booktitle={ICCV},
  pages={10895--10904},
  year={2023}
}

@software{minksolver,
    author = {Zakka, Kevin},
    title = {{Mink: Python inverse kinematics based on MuJoCo}},
    year = {2026}
}

@article{pan2025spider,
  title={Spider: Scalable physics-informed dexterous retargeting},
  author={Pan, Chaoyi and others},
  journal={arXiv preprint arXiv:2511.09484},
  year={2025}
}

@article{wampler1986manipulator,
  title={Manipulator inverse kinematic solutions based on vector formulations and damped least-squares methods},
  author={Wampler, Charles W},
  journal={IEEE Transactions on Systems, Man, and Cybernetics},
  volume={16},
  number={1},
  pages={93--101},
  year={1986},
  publisher={IEEE}
}

@article{zakka2026mjlab,
  title   = {mjlab: A Lightweight Framework for GPU-Accelerated Robot Learning},
  author  = {Zakka, Kevin and Liao, Qiayuan and Yi, Brent and Lay, Louis Le and Sreenath, Koushil and Abbeel, Pieter},
  journal = {arXiv preprint arXiv:2601.22074},
  year    = {2026}
}

\end{document}